\documentclass[lettersize,journal]{IEEEtran}
\usepackage{amsmath,amsfonts}
\usepackage{algorithmic}
\usepackage{array}
\usepackage{textcomp}
\usepackage{stfloats}
\usepackage{url}
\usepackage{verbatim}

\usepackage{graphicx}
\usepackage{booktabs}
\usepackage{threeparttable}
\usepackage{multirow}
\usepackage{amssymb}
\usepackage{makecell}
\usepackage[caption=false,font=footnotesize,labelfont=rm,textfont=rm]{subfig}

\usepackage{color}

\def\BibTeX{{\rm B\kern-.05em{\sc i\kern-.025em b}\kern-.08em
    T\kern-.1667em\lower.7ex\hbox{E}\kern-.125emX}}
\usepackage{balance}
\begin{document}
\title{LPN: Language-guided Prototypical Network for few-shot classification}

\author{Kaihui Cheng, Chule Yang*, Xiao Liu, Naiyang Guan, and Zhiyuan Wang
	\thanks{*Corresponding author}
	\thanks{Kaihui Cheng, Chule Yang , Xiao Liu, Naiyang Guan, and Zhiyuan Wang are with the Defense Innovation Institute (DII), Academy of Military Sciences, Beijing 100071, China (e-mail: chengkaihui1999@126.com, yangchule@126.com)}
        \thanks{*This work was supported by the National Natural Science Foundation of China under Grants 62006242 and 62106258. (Corresponding author: Chule Yang)}
}

\markboth{Journal of \LaTeX\ Class Files,~Vol.~18, No.~9, September~2020}%
{How to Use the IEEEtran \LaTeX \ Templates}

\maketitle

\begin{abstract}
Few-shot classification aims to adapt to new tasks with limited labeled examples. To fully use the accessible data, recent methods explore suitable measures for the similarity between the query and support images and better high-dimensional features with meta-training and pre-training strategies. However, the potential of multi-modality information has barely been explored, which may bring promising improvement for few-shot classification. In this paper, we propose a \textbf{L}anguage-guided \textbf{P}rototypical \textbf{N}etwork (LPN) for few-shot classification, which leverages the complementarity of vision and language modalities via two parallel branches to improve the classifier. Concretely, to introduce language modality with limited samples in the visual task, we leverage a pre-trained text encoder to extract class-level text features directly from class names while processing images with a conventional image encoder. Then, we introduce a language-guided decoder to obtain text features corresponding to each image by aligning class-level features with visual features.   Additionally, we utilize class-level features and prototypes to build a refined prototypical head, which generates robust prototypes in the text branch for follow-up measurement. Furthermore, we leverage the class-level features to align the visual features, capturing more class-relevant visual features. Finally, we aggregate the visual and text logits to calibrate the deviation of a single modality, enhancing the overall performance.  Extensive experiments demonstrate the competitiveness of LPN against state-of-the-art methods on benchmark datasets. 
\end{abstract}


\section{Introduction}\label{sec:Intro}
Image classification is one of the most basic computer vision tasks, which could be applied to many fields, such as face recognition ~\cite{schroff2015facenet,meng2021magface} and image understanding~\cite{russakovsky2015imagenet}. Deep neural network-based methods have made remarkable progress in this task in recent years. However, these methods require a large amount of labeled data for training, which is time-consuming to collect and costly to annotate. The models also struggle to adapt to new tasks with different data distributions from the training datasets. Unlike conventional image classification, few-shot classification aims to handle new classification tasks with limited labeled training data. However, this challenges the model to learn generalizable features from a few examples, which tends to result in overfitting. 

To address this challenge, previous studies have adopted the episode mechanism~\cite{vinyals2016matching} in few-shot classification. This approach constructs N-way K-shot learning tasks from the training data to simulate the target tasks, where each task contains a support set with labeled samples and a query set with unlabeled samples. The goal of episode methods is to learn knowledge by imitating the target task and generalizing it to new tasks with new classes. 

The few-shot classification methods can be broadly categorized into fine-tuning, meta-learning, and metric-based methods. Fine-tuning methods, such as those proposed in~\cite{chen2019closer,rajasegaran2020self}, first pre-train models on a large dataset and then fine-tune them on limited labeled data. Meta-learning~\cite{finn2017model,raghu2019rapid,raghu2019rapid} learns meta-knowledge from data and task adaptation, allowing models to adapt to new tasks rapidly. Metric-based methods, such as those presented in \cite{snell2017prototypical,li2019revisiting,sung2018learning}, aim to learn to compare the similarity between input queries and support sets with episodic training strategies. 

 %介绍一下few-shot的一些方法

Among these methods, the metric-based methods have been wildly considered. These methods rely on the fact that neural networks typically produce similar responses for objects in the same class, resulting in comparable feature representations in feature space. The key idea of metric-based methods is to leverage these characteristics to make predictions.
Despite the remarkable progress made in few-shot classification, the potential of multi-modality complementarity remains to be further explored. Humans rely on textual descriptions and visual information to acquire influential semantic information and grasp new concepts. Similarly, deep neural networks depend on high-dimensional semantic features to handle various tasks, especially when labeled data is limited. Therefore, enhancing semantic information is a promising research direction for improving few-shot classification performance. Some recent works~\cite{xing2019adaptive,yang_wang_chen_2023} have explored the cooperation of vision and language modalities for few-shot classification, demonstrating that additional semantic features, such as class names, can benefit few-shot classifiers. However, these methods leverage weighting factors based on text descriptions to adjust visual prototypes or aligning the visual features. This kind of accommodate or single post-fusion may not be adaptable to new tasks and fail to fully exploit the complementary of multi-modality. Nevertheless, these methods modify visual prototypes or align visual features based on the text features. While this unilateral alignment and singular post-fusion approach have shown promising results, they may exhibit limitations in terms of adaptability to new tasks and fail to fully exploit the inherent complementarity offered by multi-modality.

% need paired image-text descriptions
%简单介绍一下LPN的paradigm

\begin{figure*}[!t]
\centering
\includegraphics[width=1.0\linewidth]{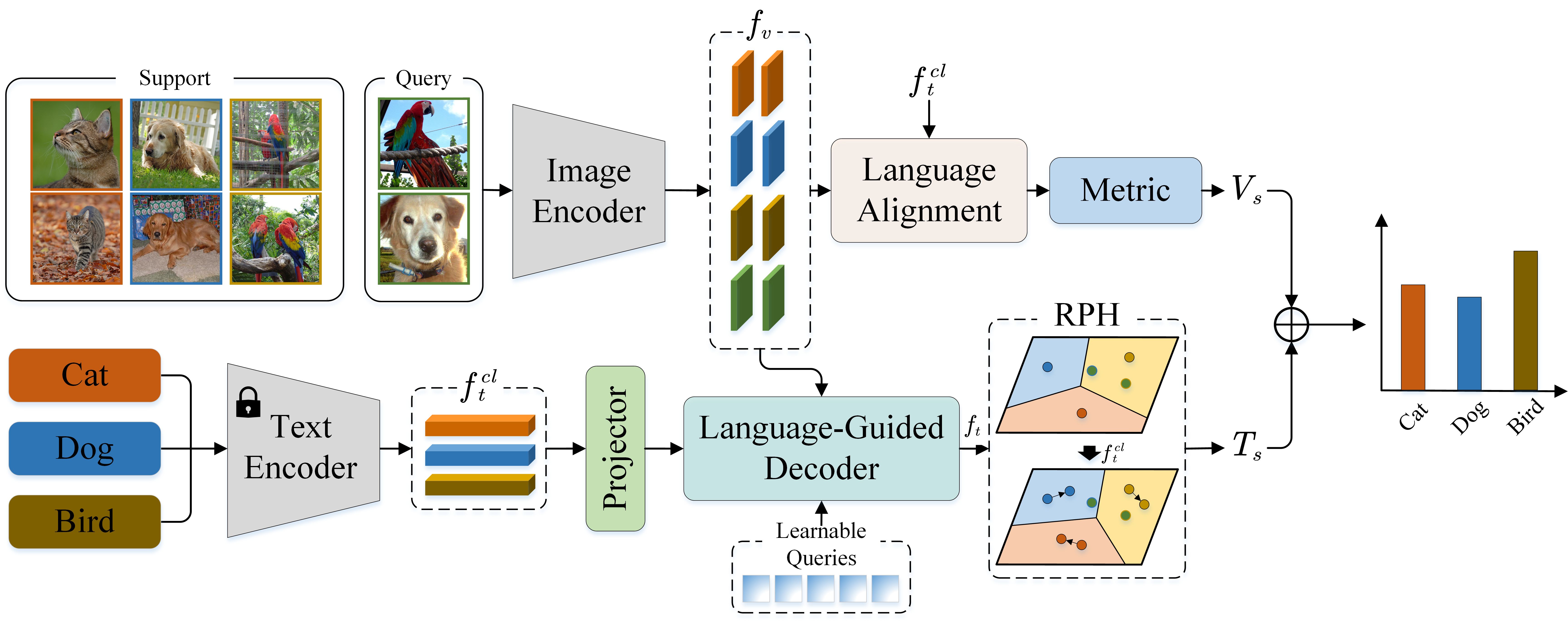}
\caption{Illustration of LPN for three-way two-shot few-shot classification. Given the task, we generate feature maps $f_v$ with the image encoder and build visual logits $V_s$ through a conventional metric module. As for the language modality, we leverage the pre-trained text encoder to extract class-level text features $f_t^{cl}$. Then, we propose a language-guided decoder to obtain the corresponding text features. LPN computes the text logits $T_s$ by a refined prototypical head, which leverages $f_t^{cl}$ to tweak the prototypes. Finally, we aggregate $V_s$ and $T_s$ to calibrate the two modalities.}
\label{fig:framework}
\end{figure*}

To address the above challenges, we propose a language-guided prototypical network (LPN) for few-shot classification that integrates vision and language modalities in the unified feature space. The two modalities provide different perspectives on the same concept and help reduce the final decision bias. However, unlike the previous multi-modality methods, we leverage both pre-fusion and post-fusion methods to take advantage of the complementarity between the two modalities and reduce decision bias. As shown in Figure~\ref{fig:framework}, LPN consists of two parallel pipelines: one for the visual branch and one for the language branch. 
The visual branch is constructed from metric-based methods. We leverage an image encoder to extra visual features and the visual features with class-level textual features, and then compute the cosine similarity between supports and queries to obtain the visual logits.
In the text branch, we first project the class names to obtain class-level text features via a pra-trained text encoder. As text descriptions are unavailable for every image in the datasets, we propose a language-guided decoder that integrates the class-level  features with visual features to generate the corresponding text features for each image (\emph{i.e.} pre-fusion). Then a refined prototypical head is used to obtain the logits of the text branches, which combines the computed prototype with class-level features. We calibrates the decision deviation by aggregating the two logits (\emph{i.e.} post-fusion). Moreover, we further constrain the generated text features through a supervised contrastive loss to obtain more distinguishable features. Our main contributions could be summarized as follows:
\begin{itemize}
    \item We present a language-guided prototypical network (LPN) for few-shot classification, which uses the complementarity of vision and language modalities to boost the metric-based classifier.
    \item We propose a language-guided decoder to transfer the knowledge from the pre-train text encoder and learn text features for each image aided by the learnable queries.
    \item A refined prototypical head is introduced to refine the prototypes with the class-level text features.
    \item Extensive experiments demonstrate the effectiveness of our LPN, and our method can generalize to other metric-based few-shot classifiers.
\end{itemize}

% The rest of this paper is organized as follows: In Section~\ref{sec:RELATEDWORK}, we review the related literature. In Section~\ref{sec:METHODOLOGY}, we present our method in detail. We report and analyze the experimental results in Section~\ref{sec:EXPERIMENTS}. Finally, we conclude in Section~\ref{sec:CONCLUSION}.

\section{RELATEDWORK}\label{sec:RELATEDWORK}

\subsection{Few-shot Classification}

Deep neural networks have achieved significant success in many fields~\cite{he2016deep,girshick2015fast,ronneberger2015u}. However, they require a large amount of labeled data, which may not be available in some scenarios. Therefore, many studies have focused on the problem of few-shot classification, which completes visual classification with limited labeled data. The fine-tuning-based methods~\cite{chen2019closer,rajasegaran2020self,liu2020negative,tian2020rethinking} first pre-train the model in the way as conventional classification or using distillation and then fine-tuning on valid labeled samples. The meta-learning-based methods~\cite{finn2017model,lee2019meta,rusu2018meta,raghu2019rapid} leverage a meta-training paradigm with a group of tasks to obtain promising initialization weights, which enables the model quickly adapt to new tasks. Besides, the metric-based methods~\cite{snell2017prototypical,li2019revisiting,sung2018learning,zhang2020deepemd} have attracted considerable research interest. They compare the similarity between the query and supports during the episodic training. 

Concretely, for the metric-based learning methods, unlabeled queries and labeled supports are encoded into the features simultaneously. The relationship between one query and support features is utilized for classification. In this way, models could classify the category of queries through comparison and quickly adapt to new tasks.  The prototypical networks (ProtoNet)~\cite{snell2017prototypical}  leverages the mean vector of global features as the prototype vector for each class, then calculates the Euclidean or cosine distance between queries and prototypes for each class to obtain the posterior probability distributions of queries. Since the global average pooling operation may degrade discriminative local features, DN4~\cite{li2019revisiting} directly finds top-k nearest neighbor support features for each query. It obtains the image-to-class similarity by matching features. Unlike the above regular metric function, the relational network~\cite{sung2018learning} introduces a learnable nonlinear metric to generate image-to-class similarity. 
% FRN~\cite{wertheimer2021few} constructs query feature maps from supports using ridge regression and computes the distance between the optimal reconstruction and the original. Besides, IEPT~\cite{zhang2021iept} generates pseudo tasks with geometric transformation to expand the support set and trains models with self-supervised learning.
BML~\cite{zhou2021binocular} introduces a binocular mutual learning paradigm to aggregate global and local features with an elastic loss. While we share the same measurement metric as the metric-based methods, we propose a language-guided prototypical network to leverage the complementarity of multi-modality to improve the few-shot classification.

\subsection{Multi-Modality in Few-shot Classification}

Few-shot classification is a challenging problem that requires flexible classifiers to learn from limited information and adapt to new tasks. To address this challenge, some works~\cite{wang2020large,xing2019adaptive,pahde2021multimodal,schwartz2022baby,zhang2021dizygotic} explore other modalities to enhance the feature extraction ability of models in a few-shot classification. In general, multi-modality few-shot learning tries to take advantage of the complementarity of modalities for better performance. Wang et al.~\cite{wang2020large} introduce weak supervision to implicit textual knowledge. Based on the prototype network, Xing et al.~\cite{xing2019adaptive} propose an adaptive modality mixture model(AM3) to adaptively adjust the prototypes through a convex combination of visual and language modalities. Schwartz et al.~\cite{schwartz2022baby} unitize multiple semantic information to boost the AM3. Yang et al.~\cite{yang_wang_chen_2022} propose a semantic guided attention mechanism (SEGA) to capture the distinguishing features. Yang et al.~\cite{yang_wang_chen_2023} leverage the combination of latent parts embeddings (LPE) from semantic knowledge to replenish the representation.  
% Chen at al.~\cite{chen2023semantic} propose a semantic prompt to adaptive tune the features through the spatial and channel interaction.

Furthermore, alongside the integration of the language modality into the primary visual extraction network, researchers have fervently pursued the incorporation of supplementary architectures to further enhance its performance.
Zhang et al.~\cite{zhang2021dizygotic} exploit two conditional variational autoencoders to adaptive combine features from different modalities. Peng et al.~\cite{peng2019few} leverage graph convolution network and knowledge graph to employ semantic information explicitly. Li et al.~\cite{li2019large} further use a class hierarchy to encode the semantic relations. In addition to feature enhancement methods for visual features, Pahde et al.~\cite{pahde2021multimodal} leverages the generative adversarial network~\cite{goodfellow2020generative}  to generate extra visual features for more reliable prototypes during training.
Similar to the previous methods, our method aggregates features from different modalities. However, instead of tuning the visual prototypes or features with the semantic knowledge, we employs parallel branches to aggregate features from different modalities. By capitalizing on the complementarity between vision and language, our method achieves a significant improvement in model performance.

\subsection{Contrastive Language-Image Pre-training}

Contrastive language-image pre-training (CLIP)~\cite{radford2021learning} is a state-of-the-art unsupervised method that learns high-quality visual features from large-scale image-text paired data. It consists of a visual encoder and a text encoder that capture the relationships between images and their corresponding texts. CLIP achieves impressive results on many visual tasks and inspires many applications~\cite{rao2022denseclip,zhou2022learning}, especially for few-shot settings. For example, CLIP-Adapter~\cite{gao2021clip} adds an extra multi-layer perceptron (MLP) to the frozen CLIP model for downstream few-shot tasks and combines outputs with language knowledge. Zhang et al. propose Tip-Adapter~\cite{zhang2021tip}, which uses a key-value cache model and updates the knowledge from CLIP for few-shot classification. In this work, we leverage the pre-trained CLIP only for the text encoder to extract class-level text features without additional training parameters.

%%%%%%%%%%%%%%%%%%%%%%%%%%%%%%%%%%%%%%%%%%%%%%%%%%%%%%%%%%%%%%

\section{METHODOLOGY}\label{sec:METHODOLOGY}
\subsection{Preliminary}
In the standard few-shot classification, it consists of a base dataset $\mathcal{C}_{base}$ for training and a novel dataset $\mathcal{C}_{novel}$ for evaluation, where $\mathcal{C}_{base}\cap \mathcal{C}_{novel} = \varnothing$. During training, the episodic-train strategy is usually adopted to construct a set of tasks $\left \{ \mathcal{T} \right \}_1^n$ to simulate the target scenario. In the same measure, evaluation is performed on the tasks $\left \{ \mathcal{T} \right \}_1^n$. The models are supposed to learn classifiers from $\mathcal{C}_{base}$ that could quickly adapt to novel tasks built from $\mathcal{C}_{novel}$ given a few labeled samples (\emph{i.e.} supports). Specifically, each task $\mathcal{T}$ contains a support set  $\mathcal{S}$ and a query set $\mathcal{Q}$. For a $N$-way $K$-shot few-shot classification task, the support set $\mathcal{S}$ has $N$ classes, and each class consists of $K$ labeled images. the query set $\mathcal{Q}$ involves $N\times Q$ images without labels.

\subsection{Method Overview}

The overall illustration of our method is given in Figure~\ref{fig:framework}. Language-Guided Prototypical Networks (LPN) utilize the knowledge from language modality to boost the performance of few-shot classification. LPN comprises two branches: the visual branch and the text branch. The visual branch extracts feature $f_v$ from the input image and aligns features with the class-level features, then we could obtain classification logits $V_s$ through the metric module. The text branch utilizes a pre-trained text encoder, coupled with a projector to generate class-level text features $f_t^{cl}$ from the category names in the task. Based on the transformer architecture, we propose a language-guided decoder to integrate $f_t^{cl}$ with visual features $f_v$ via the learnable queries. Thus, we could obtain the text features corresponding to each image in the pre-fusion. The text branch also employs a refined prototypical head to improve the assessment of prototypes and then generate classification logits $T_s$. In the post-fusion stage, we aggregate $V_s$ and $T_s$ to calibrate the logits and obtain the predictions.

\subsection{Language Alignment}\label{sec:LA}

% Inspired by the LPE~\cite{yang_wang_chen_2023}, 
After the image encoder extracts the visual features $f_v$, we further leverage the class-level text features $f_t^{cl}$, which represent textual information associated with each category, as convolutional kernels for different categories. Concretely, for the $j$-th class, we transform the class-level features $f_t^{cl_j}$ into the convolutional kernel $\mathcal{K}_j \in \mathcal{R}^{D\times 1 \times 1}$ to match the dimensions required for the convolution operation. We apply the kernel $\mathcal{K}_j$ on the $j$-th class visual features $f_v^j$ of the supports to extract class-specific information from the visual features. To determine the relevance of the transformed visual features for each class, we use a sigmoid function to compute the attention weight, indicating the significance of the visual features in relation to the corresponding class:
\begin{equation}
\hat{f}_v^j = f_v^j \odot sigmoid (f_v^j * \mathcal{K}_j)
\label{equ:LA}
\end{equation}
where $\odot$ is Hadamard product, and $*$ refers to the convolution operation. The aligned visual features of supports $\hat{f}_v^j $ are further used to compute the similarity in the visual branch.

\begin{figure}[!t]
\flushleft
\includegraphics[width=1.0\linewidth]{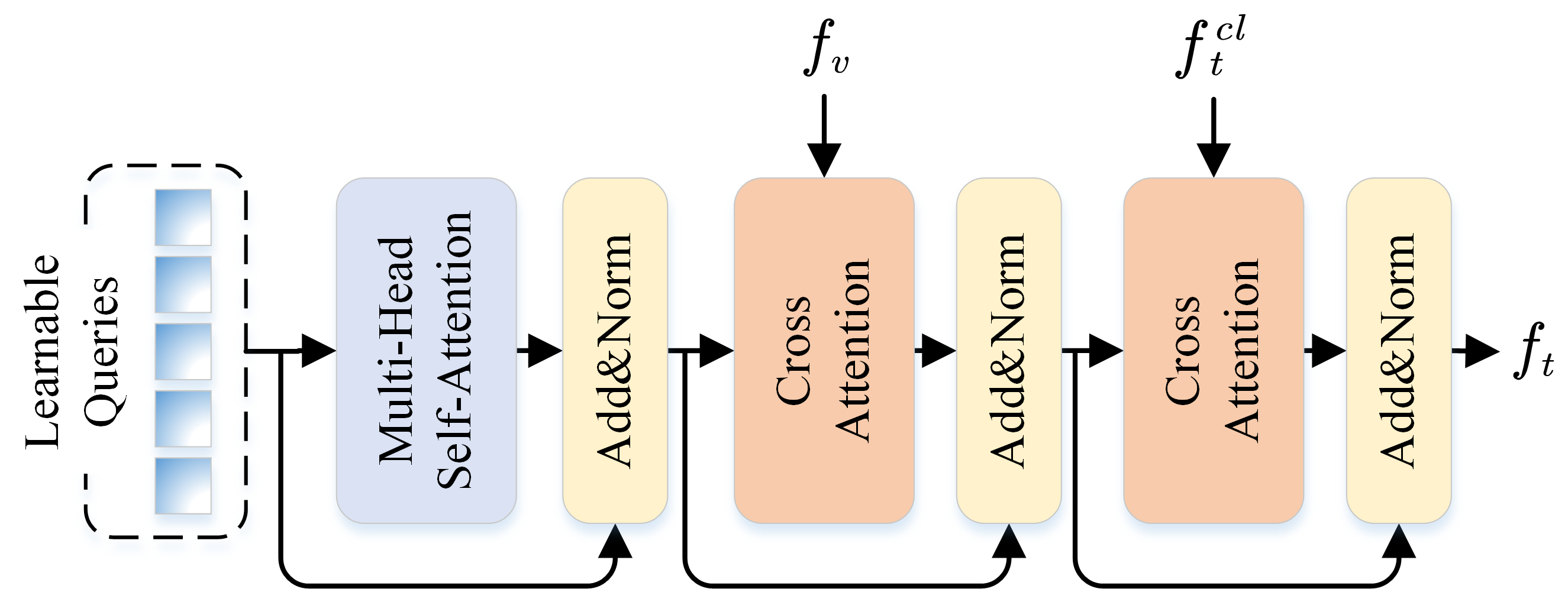}
\caption{Illustration of the language-guided decoder. The learnable queries are encoded via a multi-head self-attention, and then text features are obtained through two cross-attention modules.}
\label{fig:LGD}
\end{figure}

\subsection{Language-Guided Decoder}\label{sec:LaGD}

Obtaining text features that correspond to images directly through the CLIP text encoder is challenging due to the absence of paired text descriptions. We propose a language-guided decoder (LaGD) module for pre-fusion to address this issue. The LaGD pipeline follows the standard transformer architecture \cite{vaswani2017attention} with multi-head self-attention and cross-attention, as shown in Figure~\ref{fig:LGD}. The multi-head attention mechanism can be described as follows:
\begin{equation}
MultiHead\left( q,k,v \right) =Concat\left( \left[ h_1,...,h_n \right] \right) W_o
\label{equ:multihead}
\end{equation}
where $Concat(\cdot)$ is concatenation operation along the channels, $W_o$ is a learnable weight matrix. $h_i$ refers to $i$-th head:
\begin{equation}
h_i=Attention\left( qW_{i}^{q},kW_{i}^{k},vW_{i}^{v} \right)
\label{equ:hi}
\end{equation}
$W_{i}^{q}$,$W_{i}^{k}$ and $W_{i}^{v}$ are $i$-th learnable weight matrix used for different representation subspaces, respectively. The operation of $Attention(\cdot)$ is formulated as:
\begin{equation}
Attention\left( q,k,v \right) =softmax\left( \frac{qk^T}{\sqrt{d_k}}v \right)
\label{equ:attn}
\end{equation}
$\sqrt{d_k}$ is the dimension of head.

Our LaGD begins by improving the quality of learnable queries using multi-head self-attention. Then, we leverage cross-attention to connect these queries with respect to the visual feature maps $f_v$ from the image encoder, resulting in visual-specific queries. We extract class-level text features $f_t^{cl}$ through the text encoder to generate text features. By combining visual-specific queries and $f_t^{cl}$, we can exploit text features $f_t$.

% We start by mapping class names through the text encoder to extract ground truth text features $T_{gt}$.

\begin{figure}[!t]
\flushleft
\includegraphics[width=1.0\linewidth]{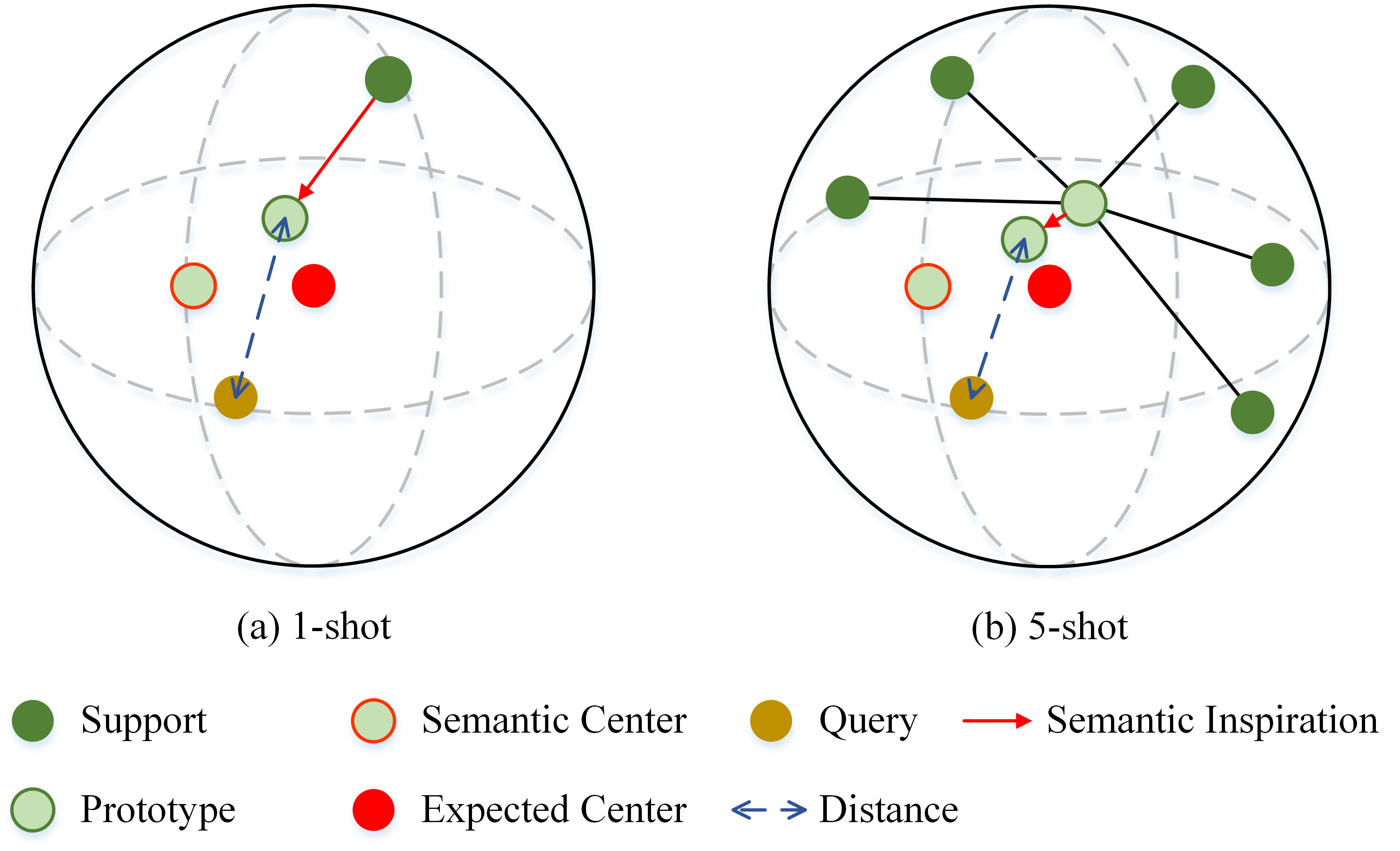}
\caption{The schematic overview of the refined prototypical head. The class-level feature is used to adjust the computed prototype.}
\label{fig:RPH}
\end{figure}

\subsection{Refined Prototypical Head} 
\label{sec:rph}

The LaGD obtains text features based on image and class-level information, and we naturally adopt a prototypical module to measure the similarity of text features in the text branch. However, the generated features may not be diverse enough to generate prototypes accurately, resulting in a gap between the generated prototype vector and the expected one. On the other hand, directly using class-level text as a potential prototype may affect the model's performance by overlooking individual characteristics critical to prototype building.

To overcome this issue, we propose a refined prototypical head incorporating class-level features, allowing for adjustments to the prototype vector and improving its overall quality, as shown in Figure~\ref{fig:RPH}. The single support text feature could be treated as the class prototype for the one-shot few-shot classification task. It may not be enough to adequately represent the class prototype due to the lack of diversity. While the class-level feature may not match the expected prototype, it can still be leveraged to align the computed prototype with the expected. We can obtain a more precise distance in the final decision by doing so.  In contrast, the five-shot task includes more labeled samples, making the prototype computed from text features closer to the expected one. In this case, only slight shaping of the prototype may be necessary.

Specifically, given the few-shot classification task $\mathcal{T}$, we leverage LaGD to obtain corresponding text features $f_{t}^i$ for each visual feature $f_{v}^i$. The text prototype $c_j$ for class $j$ could be formulated as:
\begin{equation}
c_j =\frac{1}{K}\sum _{i=1}^K f_{t}^i
\label{equ:prototype}
\end{equation}
where $c_j\in \mathbb{R}^d$, $d$ is the dimension of text feature, $K$ indicates $K$ images from the support set $\mathcal{S}$ for class $j$. The refined prototype is,
\begin{equation}
RPH(c_j) =\gamma c_j +(1-\gamma)f_t^{cl_j}
\label{equ:RPH}
\end{equation}
$RPH(\cdot)$ is the refined prototypical head, $\gamma$ is a hyperparameter, $f_t^{cl_j}$ refers to the $j$-th class-level text features generated from the class name via the text encoder.

\subsection{Loss Function}
\label{sec:loss}
We leverage the metric-based methods to obtain the logits. Given a few-shot task and the metric, the visual logits $V_s$ could be calculated as:
\begin{equation}
    V_s=Metric(\mathcal{S},\mathcal{Q})
\label{equ:Vscore}
\end{equation}
where the $Metric(\cdot)$ refers to these conventional measurement, such as prototype~\cite{snell2017prototypical}, local descriptions~\cite{li2019revisiting}, \emph{etc}. The text branch utilizes RPH and cosine similarity for measurement. The text logits for $i$-th query sample are:
\begin{equation}
    T_s=\frac{f_t^i \cdot c_j}{\Vert f_t^i \Vert_2 \cdot \Vert c_j \Vert_2}
\label{equ:Tscore}
\end{equation}
We further integrate the logits of visual and text branches and final logits $\mathbf{s}=V_s+T_s$. Thus the posterior probability distribution of a query sample can be summarised as follows:
\begin{equation}
    \rho(y=k|Q_i)= \frac{\exp(\alpha \cdot \mathbf{s}_k)}{\sum_{j=1}^C\exp(\alpha \cdot \mathbf{s}_j)}
\label{equ:prob}
\end{equation}
where $\alpha$ is a scaler factor. The standard cross-entropy loss $\mathcal{L}_c$  is used for classification.

As we generate class-level text features directly from the class names, there may be a risk of confusion between these features. We utilize a supervised contrastive loss function $\mathcal{L}_{scl}$ to modify the text features to alleviate this. This differs from prompt engine methods such as those used in \cite{radford2021learning, zhou2022learning}.
\begin{small}  %设置字体大小  可以选择 small  tiny
 \begin{equation}
        \mathcal{L}_{scl}=-\frac{1}{M}\sum _{i=1}^{M}\log\frac{\exp(f_t^i\cdot g_{i}^{+}/\tau)}{\exp(f_t^i\cdot g_{i}^{+}/\tau)+\sum_{j=1,j\neq i}^{N}\exp(f_t^i\cdot g_{j}^{-}/\tau)}
    \label{equ:Lscl}
\end{equation}
\end{small}%
where $f_t^i$ is the $i$-th text features, $g_{i}^{+}$ is the positive features, and $g_{j}^{-}$ represents the negative features. $\tau$ is the temperature parameter, and $M$ is the batch's total number of text features. We combine the two losses with a weight factor $\beta$ as a total loss.
 \begin{equation}
    \mathcal{L}=\mathcal{L}_c+\beta \mathcal{L}_{scl}
\label{equ:L}
\end{equation}
%%%%%%%%%%%%%%%%%%%%%%%%%%%%%%%%%%%%%%%%%%%%%%%%%%%%%%%%%%%%%%

\begin{table*}[!t]
    \centering
    \caption{The 5-way 1(5)-shot classification results(\%) on \emph{mini}ImageNet, \emph{tiered}ImageNet. The notation $^\sharp$ refers that the results reported in \cite{lee2019meta}. Language indicates whether the language modality is used.}
    \label{Tab:ImageNetserious}
    \begin{tabular}{lcccccc}
    \toprule
    \multirow{2}{*}{\textbf{Method}}&\multirow{2}{*}{\textbf{Backbone}}&\multirow{2}{*}{\textbf{Language}}&\multicolumn{2}{c}{\textbf{\emph{mini}ImageNet}}&\multicolumn{2}{c}{\textbf{\emph{tiered}ImageNet}}
    \\ \cline{4-7}
    &&&\emph{\textbf{1-shot}}&\emph{\textbf{5-shot}}&\emph{\textbf{1-shot}}&\emph{\textbf{5-shot}}
    \\ \midrule
    \textbf{MAML}$^\sharp$~\cite{finn2017model}&\emph{ConvNet}&No&48.70$\pm$1.84&63.11$\pm$0.92&51.67$\pm$1.81&70.30$\pm$1.75\\ 
    \textbf{ProtoNet}$^\sharp$~\cite{snell2017prototypical}&\emph{ConvNet}&No&49.42$\pm$0.78&68.20$\pm$0.66&53.31$\pm$0.89&72.69$\pm$0.74\\ 
    \textbf{MatchingNet}$^\sharp$~\cite{vinyals2016matching}&\emph{ConvNet}&No&43.56$\pm$0.84&55.31$\pm$0.73&-&-\\ 
    \textbf{RelationNet}$^\sharp$~\cite{sung2018learning}&\emph{ConvNet}&No&50.44$\pm$0.82&65.32$\pm$0.70&54.48$\pm$0.93&71.32$\pm$0.78\\ 
    \midrule
    \textbf{CTM}~\cite{li2019finding}&\emph{ResNet-18}&No&64.12$\pm$0.82&80.51$\pm$0.13&68.41$\pm$0.39&84.28$\pm$1.73\\
    \textbf{Centroid}~\cite{afrasiyabi2020associative}&\emph{ResNet-18}&No&59.88$\pm$0.67&80.35$\pm$0.73&69.29$\pm$0.56&85.97$\pm$0.49\\

    \textbf{TADAM}~\cite{oreshkin2018tadam}&\emph{ResNet-12}&No&58.50$\pm$0.30&76.70$\pm$0.30&-&-\\ 
    \textbf{MetaOptNet}~\cite{lee2019meta}&\emph{ResNet-12}&No&64.09$\pm$0.62&80.00$\pm$0.45&65.81$\pm$0.74&81.75$\pm$0.53 \\
    
    \textbf{FEAT}~\cite{ye2020few}&\emph{ResNet-12}&No&66.78$\pm$0.20&82.05$\pm$0.14&70.80$\pm$0.23&84.79$\pm$0.16\\ 
    
    \textbf{RFS}~\cite{tian2020rethinking}&\emph{ResNet-12}&No&64.82$\pm$0.60&82.14$\pm$0.43&71.52$\pm$0.69&86.03$\pm$0.49\\
    \textbf{Meta-Baseline}~\cite{chen2020new}&\emph{ResNet-12}&No&63.17$\pm$0.23&79.26$\pm$0.17&68.62$\pm$0.27&83.29$\pm$0.18\\
    \textbf{Neg-Cosine}~\cite{liu2020negative}&\emph{ResNet-12}&No&63.85$\pm$0.81&81.57$\pm$0.56&-&-\\
    \textbf{DeepEMD}~\cite{zhang2020deepemd}&\emph{ResNet-12}&No&65.91$\pm$0.82&82.41$\pm$0.56&71.16$\pm$0.87&86.03$\pm$0.58\\ 
     \textbf{FRN}~\cite{wertheimer2021few}&\emph{ResNet-12}&No&66.45$\pm$0.19&82.83$\pm$0.13&71.16$\pm$0.22&86.01$\pm$0.15\\
     \textbf{BML}~\cite{zhou2021binocular}&\emph{ResNet-12}&No&67.04$\pm$0.63&\textbf{83.63$\pm$0.29}&68.99$\pm$0.50&85.49$\pm$0.34\\
     % \textbf{IEPT}~\cite{zhang2021iept}&\emph{ResNet-12}&67.05$\pm$0.44&82.90$\pm$0.30&72.24$\pm$0.50&\textbf{86.73$\pm$0.34}\\
    \midrule 
    \textbf{AM3}~\cite{xing2019adaptive}&\emph{ResNet-12}&Yes&65.21$\pm$0.49&75.20$\pm$0.27&67.23$\pm$0.34&78.95$\pm$0.22\\
    \textbf{SEGA}~\cite{yang_wang_chen_2022}&\emph{ResNet-12}&Yes & 69.04$\pm$0.26 & 79.03$\pm$0.18 & 72.18$\pm$0.30 & 84.28$\pm$0.21 \\
    \textbf{LPE}~\cite{yang_wang_chen_2023}&\emph{ResNet-12}&Yes & 71.64$\pm$0.40 & 79.67$\pm$0.32 & 73.88$\pm$0.48 & 84.88$\pm$ 0.36\\
    % \textbf{SP}~\cite{chen2023semantic}&\emph{Visformer-T} & $\pm$ & $\pm$ & $\pm$ & $\pm$ \\
    % \textbf{Baseline(\emph{ProtoNet})}~\cite{snell2017prototypical}&\emph{ResNet-12}&63.33$\pm$0.45&80.81$\pm$0.30&70.81$\pm$0.50&85.75$\pm$0.34\\ 
    % \textbf{MatchingNet}$^\dagger$~\cite{vinyals2016matching}&\emph{ResNet-12}&60.47$\pm$0.43&77.13$\pm$0.32&-&-\\ 
    % \textbf{DN4}$^\dagger$~\cite{li2019revisiting}&\emph{ResNet-12}&61.31$\pm$0.45&76.02$\pm$0.35&-&-\\ 
    \midrule 
    % \textbf{LPN-old}&\emph{ResNet-12}&Yes&\textbf{68.04$\pm$0.42}&82.15$\pm$0.30&\textbf{74.85$\pm$0.48}&86.07$\pm$0.34\\ 
    \textbf{LPN}&\emph{ResNet-12}&Yes&\textbf{71.99$\pm$0.38} & 82.43$\pm$0.30& \textbf{76.77$\pm$0.46}&\textbf{86.09$\pm$0.34}\\ 
    % \textbf{LPN-V}&\emph{Visformer-T}& $\pm$ & $\pm$ & $\pm$ & $\pm$ \\
    \bottomrule
    \end{tabular}
\end{table*}

%%%%%%%%%%%%%%%%%%%%%%%%%%%%%%%%%%%%%%%%%%%%%%%%%%%%%%%%%%%%%%
\section{EXPERIMENTS}\label{sec:EXPERIMENTS}
% The language-Guided prototypical network leverages the complementary nature of vision and language to improve the efficiency of classifiers in few-shot settings. We demonstrate the superiority and generality of LPN on standard benchmark datasets.

\subsection{Datasets}

We perform the experiments on three widely-used benchmark datasets: \emph{mini}ImageNet~\cite{vinyals2016matching}, \emph{tiered}ImageNet~\cite{ren2018meta}, CUB-200-2011(CUB)~\cite{wah2011caltech}. \emph{mini}ImageNet contains 100 classes with 600 images per class sampled from the ILSVRC-2012~\cite{russakovsky2015imagenet}, and we use 64/16/20 classes for train/val/test, respectively. \emph{tiered}ImageNet is a larger version of \emph{mini}ImageNet with 608 classes and 779,165 images, and we use 351/97/160 classes for train/val/test, respectively. CUB is a fine-grained classification dataset with 200 bird species and 11,788 images, and we use 100/50/50 classes for train/val/test, respectively.

\subsection{Implement Details} \label{sec:implement}

% \noindent
\subsubsection{Pipeline}
The experiments are conducted on ResNet-12 and ResNet-18~\cite{he2016deep}, each containing four residual blocks. The input resolution for ResNet-12 is 84$\times$84. The number of filters is set to (64, 160, 320, 640) for ResNet-12. The number of attention heads is set to 16. We utilize a linear projection to align the channel of text and visual features between the two cross-attention modules in the language-guided decoder. The projector contains 2 fully connected layers with ReLU activation function between them. The pre-trained CLIP (RN50) is used as our default text encoder. 

We train the LPN using a meta-learning framework with episodic training tasks. Each episodic task consists of standard 5-way 1-shot or 5-way 5-shot tasks sampled from the training data. Before episodic training, we pre-train our models on the training data and use the resulting weights for initialization, following the approach used in~\cite{chen2021meta}. All our few-shot classification experiments are performed with prototypes~\cite{snell2017prototypical} as the visual metric unless otherwise stated.

% \noindent 
\subsubsection{Optimization}

In our experimental setup, we leverage the ResNet-12 architecture as the backbone and the batch size is fixed to 64. In the pre-training stage, we employ an SGD optimizer with a learning rate of 5e-2, weight decayof 1e-4, momentum of 0.9, over 200 epochs for mini-ImageNet and CUB. The learning rate is decreased by a factor of 1/10 at epochs 100 and 150. For the tiered-ImageNet dataset, we use the same optimizer with a learning rate of 5e-2, weight decay of 1e-4, momentum of 0.9, over 100 epochs. The learning rate is decreased at epochs 40 and 70. 

In the meta-training stage, we employ the SGD optimizer with a learning rate of 5e-4 for 1-shot tasks. For 5-shot tasks, except for the mini-ImageNet dataset where the learning rate is adjusted to 5e-5, the other datasets maintain a learning rate of 1e-4. Learning rate decay occurs at epochs 40 and 80 for the mini-ImageNet and CUB datasets, whereas for the tiered-ImageNet dataset, the decay takes place at epochs 40 and 70. Notably, the CUB dataset encompasses 600 episodes within each epoch, while all other datasets complete 1000 episodes. The decay factor applied is 0.1.

The scale factor, $\alpha$, is set to 10, and the hyperparameter $\beta$ in the loss function is 0.4. The RPH ratio $\gamma$ is 0.01 and 0.1 for 1-shot and 5-shot tasks, respectively. We report the mean accuracy of 2000 episodes with 95\% confidence intervals. 

%%%%%%%%%%%%%%%%%%%%%%%%%%%%%%%%%%%%%%%%%%%%%%%%%%%%%%%%%%%%%%

\subsection{Main Results}

Table~\ref{Tab:ImageNetserious} presents the results of our experiments on the coarse-grained benchmark, which demonstrate that LPN achieves competitive performance in few-shot learning tasks, particularly in 1-shot settings. The filter numbers of the four blocks in the \emph{ConvNet} are set as ~\cite{lee2019meta}.

% \noindent
\subsubsection{Coarse-grained Benchmark}

In comparison to the metric-based approach DeepEMD~\cite{zhang2020deepemd}, our LPN demonstrates noteworthy enhancements in 1-shot tasks, achieving gains of 6.08\% and 5.61\%, all the while maintaining commendable performance in 5-shot tasks. As for the state-of-the-art metric-based method FRN~\cite{wertheimer2021few}, it records success rates of 66.45\% and 71.16\% in 1-shot scenarios for \emph{mini}ImageNet and \emph{tiered}ImageNet  respectively. FRN leverage the  closed form produces from reconstruction problem to measure the similarity between query and supports, in contrast, our LPN achieves substantial progress with remarkable improvements of 1.99\% and 3.69\% by incorporating the language modality and using the conventional prototype metric.  The language-based methods encounter challenges in 5-shot tasks due to the heightened stability and precision of visual embeddings, coupled with the influx of supplementary visual data, as elucidated by Yang et al.~\cite{yang_wang_chen_2023}. Compared to analogous multimodal learning techniques such as LPE~\cite{yang_wang_chen_2023}, LPN showcases significant advancements in both 1-shot and 5-shot tasks with the two parallel pipelines, which mitigates the weaken-gain problem inherent in multimodal few-shot methods as the number of supports increases.

\begin{table}[h]
    \centering
    \caption{The 5-way 1(5)-shot classification results(\%) on CUB. The notation $^\sharp$ indicates results reported in \cite{lee2019meta}.}
    \vspace{2pt}
    \label{Tab:CUB}
    % \small
    % \scalebox{0.9}{
    \begin{tabular}{lccc}
    \toprule
    \multirow{2}{*}{Method}&\multirow{2}{*}{Backbone}&\multicolumn{2}{c}{CUB}
    \\ \cline{3-4}
    &&\emph{1-shot}&\emph{5-shot}
    \\ \midrule
    % \textbf{CloserLook++}~\cite{chen2019closer}&\emph{ResNet-18}&67.02$\pm$0.90&83.58$\pm$0.54\\
    \textbf{MAML$^\sharp$}~\cite{finn2017model}&\emph{ResNet-18}&68.42$\pm$1.07&83.47$\pm$0.62\\ 
    \textbf{ProtoNet$^\sharp$}~\cite{snell2017prototypical}&\emph{ResNet-18}&72.99$\pm$0.88&86.64$\pm$0.51\\
    \textbf{MatchingNet$^\sharp$}~\cite{vinyals2016matching}&\emph{ResNet-18}&73.49$\pm$0.89&84.45$\pm$0.58\\
    \textbf{RelationNet$^\sharp$}~\cite{sung2018learning}&\emph{ResNet-18}&68.58$\pm$0.94&84.05$\pm$0.56\\
    \textbf{LaplacianShot}~\cite{ziko2020laplacian}&\emph{ResNet-18}&80.96$\pm$N/A &88.68$\pm$N/A\\
    \textbf{S2M2}~\cite{mangla2020charting}&\emph{ResNet-18}&71.43$\pm$0.28&85.55$\pm$0.52\\
    \textbf{Neg-Cosine}~\cite{liu2020negative}&\emph{ResNet-18}&72.66$\pm$0.85&89.40$\pm$0.43\\
    \textbf{Centroid}~\cite{afrasiyabi2020associative}&\emph{ResNet-18}&74.22$\pm$1.09&88.65$\pm$0.55\\
    \textbf{BML}~\cite{zhou2021binocular}&\emph{ResNet-12}&76.21$\pm$0.63&90.45$\pm$0.36\\
    \textbf{FRN}~\cite{wertheimer2021few}&\emph{ResNet-18}&83.55$\pm$0.19&\textbf{92.92$\pm$0.10}\\
    \midrule 
    \textbf{SEGA}~\cite{yang_wang_chen_2022}&\emph{ResNet-12} & 84.57$\pm$0.22 & 90.85$\pm$0.16  \\
    
    \textbf{LPE}~\cite{yang_wang_chen_2023}&\emph{ResNet-12} & 80.76$\pm$0.40 & 88.98$\pm$0.26 \\
    \textbf{LPE-attributes}~\cite{yang_wang_chen_2023}&\emph{ResNet-12} & 85.04$\pm$0.34  & 89.24$\pm$0.26 \\
    
    \midrule
    % \textbf{Baseline(\emph{ProtoNet})}~\cite{snell2017prototypical}&\emph{ResNet-18}&-$\pm$-&-$\pm$-\\ 
    % \textbf{LPN-old}&\emph{ResNet-18}&\textbf{84.74$\pm$0.35} &91.48$\pm$0.21\\ 
    % \textbf{LPN}&\emph{ResNet-18}&\textbf{85.83$\pm$0.34} & 90.86$\pm$0.22\\ 
    \textbf{LPN}&\emph{ResNet-12}& 85.78$\pm$0.33 & 91.95$\pm$0.21 \\
    \bottomrule
    \end{tabular}
    % }
\end{table}

\subsubsection{Fine-grained Benchmark}
 We also evaluate the model on the fine-grained CUB benchmark, and the results are summarized in Table~\ref{Tab:CUB}. Although we find that LPN is inferior to FRN in 5-shot tasks as it adopts naive prototypes for classification, our model outperforms the other method across the board in 1-shot tasks, which proves that the class-level text feature is efficient in training a robust classifier with minimal labeled data.

The experiments indicate that our LPN is superior in 1-shot tasks and can still achieve competitive performance in 5-shot tasks compared with other language-based methods. LPN achieves this performance without relying on well-designed metrics used by other metric-based methods, instead utilizing the conventional prototype metric and taking advantage of language modality. 
Our proposed method, LPN, differs from AM3~\cite{xing2019adaptive} and LPE~\cite{yang_wang_chen_2023} in that we introduce the text branch to handle language modality combined with the visual branch for final decision-making. This approach allows LPN to effectively retain information from both modalities and better utilize their complementarity in the post-fusion stage. In contrast, AM3 directly leverages class-level features to adjust visual prototypes without explicitly considering language modality, and LPE uses the class-level feature vector to highlight the visual features. 

By processing the features of different modalities through separate branches, LPN can better utilize the strengths of each modality and improve the overall performance of few-shot classification models.

%%%%%%%%%%%%%%%%%%%%%%%%%%%%%%%%%%%%%%%%%%%%%%%%%%%%%%%%%%%%%%%%%%%%%
\subsection{Ablation Study}

\begin{table}[!t]
    \centering
    \caption{Ablation study of proposed modules on \emph{mini}ImageNet.}
    \vspace{2pt}
    \label{Tab:albation}

    \begin{tabular}{c c c c c }%
	\toprule
		\multicolumn{3}{c}{Module} & \multirow{2}{*}{1-shot}& \multirow{2}{*}{5-shot}\\
		\cmidrule(lr){1-3}
	 LaGD& RPH &LA \\ 

	\midrule
	$\times$&$\times$&$\times$&  63.54$\pm$0.45	& 80.79$\pm$0.30 \\
	$\checkmark$&$\times$&$\times$&  64.79$\pm$0.45	& 81.70$\pm$0.30\\ 
	$\checkmark$&$\checkmark$&$\times$&  71.95$\pm$0.37 & 	81.73$\pm$0.31 \\
	$\checkmark$&$\checkmark$&$\checkmark$ & 71.99$\pm$0.38 & 	82.43$\pm$0.30  \\
    \bottomrule
\end{tabular}
\end{table}

\subsubsection{Ablation of Proposed Modules}

We conduct several experiments on \emph{mini}ImageNet to explore the effectiveness of proposed modules. The results are shown on Table~\ref{Tab:albation}. Through the incorporation of the language-guided decoder, the accuracy of the baseline model is improved by 1.25\% and 0.91\%, respectively. Leveraging the refined prototypical head to adjust the text prototypes, especially in 1-shot tasks, yields a notable improvement of 7.16\%. The results underscore the efficacy of the language alignment module, particularly in the context of 5-shot tasks.   By combing these modules together, LPN achieves excellent performance compared with the baseline model.

% \begin{figure}[!t]
%     \centering
%     \subfloat[1-shot]{
%         \includegraphics[width=0.45\linewidth]{Figure/ablation_on_LPN/1shot.png}
%         \label{fig:AS_LPN.1shot}
%     }  
%     % \quad
%     \subfloat[5-shot]{
% 	\includegraphics[width=0.45\linewidth]{Figure/ablation_on_LPN/5shot.png}
%         \label{fig:AS_LPN.5shot}
%     }
%     \caption{Ablation studies of text branch on multiple benchmarks.}
%     \label{fig:AS_LPN}
% \end{figure}

\begin{table}[!t]
    \centering
    \caption{Ablation studies of text branch on multiple benchmarks. PN represents the prototype network.}
    \vspace{2pt}
    \label{Tab:AS_LPN}

    \begin{tabular}{cccc}
    
    \toprule
    \multirow{2}{*}{DATA}&\multirow{2}{*}{\makecell{Text \\ Metric}}&\multicolumn{2}{c}{Shot}\\ 
    \cline{3-4}
    && \emph{1-shot} & \emph{5-shot}\\ 
    \midrule
    \multirow{2}{*}{\emph{mini}ImageNet} 
    &PN    &63.54 &	80.79 \\
    & LPN   &71.99$_{{(+8.45)}}$ &	82.43$_{{(+1.64)}}$\\ 
    \midrule

    \multirow{2}{*}{ \emph{tiered}ImageNet} 
    &PN     &70.63 &	85.66 \\
    & LPN   &76.77$_{{(+6.14)}}$ &	86.09$_{{(+0.43)}}$ \\ 
    \midrule
    
    \multirow{2}{*}{ CUB} 
    &PN     &78.47 &	91.01 \\
    & LPN   &85.78$_{{(+7.31)}}$ &	91.95$_{{(+0.94)}}$
    \\

    \bottomrule
    
    \end{tabular}
\end{table}

% \noindent
\subsubsection{Ablation on Text Branch}
In our proposed approach, LPN, we augment the model with additional text branches that combine class-level text features with visual features. To evaluate the effectiveness of LPN, we conduct ablation studies using the settings described in Section~\ref{sec:implement}. The results of these experiments are presented on Table ~\ref{Tab:AS_LPN}. The PN represents the prototypical network~\cite{snell2017prototypical}.  Our approach outperforms the PN by a significant margin, achieving improvements of 8.45\%, 6.14\%, and 7.31\% on 1-shot tasks for \emph{mini}ImageNet, \emph{tiered}ImageNet, and CUB, respectively. By leveraging visual and language modalities, our approach shows promising results in few-shot classification.

\begin{table}[!t]
    \centering
    \caption{Effect of PRH on multiple benchmarks. PH represents the prototype metric in the text branch.}
    \vspace{2pt}
    \label{Tab:EF_RPH}

    \begin{tabular}{ccc}
    
    \toprule
    DATA&Text Metric&\emph{1-shot}\\ 
    \midrule
    \multirow{2}{*}{\emph{mini}ImageNet} 
    &PH    &64.97  \\
    & RPH   &71.99$_{{(+7.02)}}$ \\ 
    \midrule

    \multirow{2}{*}{ \emph{tiered}ImageNet} 
    &PH     &69.58	 \\
    & RPH   &76.77$_{{(+7.19)}}$ 
    \\ 
    \midrule
    \multirow{2}{*}{ CUB} 
    &PH     &79.84	 \\
    & RPH   &85.78$_{{(+5.94)}}$    \\

    \bottomrule
    
    \end{tabular}
\end{table}

% \noindent
\subsubsection{Ablation on RPH}
To bridge the gap between calculated prototypes and expectations, we propose the RPH metric in the text branch. This metric capitalizes on both class-level and support text features to compute prototypes, enhancing the precision of similarity estimation.  We conduct experiments to evaluate the effectiveness of the proposed RPH, and the results are presented in Table ~\ref{Tab:EF_RPH}. We compared our approach to the prototypical head (PH) where $\gamma$ is set to 1, meaning only support features are used to construct prototypes in the text branch. The results show the significant enhancement brought about by RPH in 1-shot tasks. This suggests that RPH has the potential to mitigate the inconsistencies arising from insufficient labeled text samples. Consequently, the model becomes more adept at harnessing class-level text features for improved performance.

\begin{table}[!t]
    \centering
    \caption{The influence of $\gamma$ in the refined prototypical head on \emph{mini}ImageNet.}
    \vspace{2pt}
    \label{Tab:AS_gamma}
    \begin{tabular}{c | c c c c  }
	\toprule
		$\gamma$ & 0.01 &0.1 &0.5 &1.0\\
	\midrule
	\emph{1-shot} & 71.99$_{{(+7.02)}}$ &65.09$_{{(+0.12)}}$  &64.98$_{{(+0.01)}}$ &64.97\\

\end{tabular}
\end{table}

% \noindent
 \subsubsection{Effect of $\gamma$}
In Equation(\ref{equ:RPH}), we introduce the hyperparameter $\gamma$, which determines the contribution of class-level text features and prototypes in the text branch. Adjusting this parameter allows the model to balance the trade-off and generate refined prototypes that facilitate few-shot classification tasks. 
Larger gamma values indicate less intervention of textual information in the prototype construction.
As illustrated in Table \ref{Tab:AS_gamma}, incorporating additional class-level features leads to substantial improvements in the case of the 1-shot task.
Concretely, the results indicate that fewer labeled samples require more intervention of class-level features to obtain superior prototype vectors. This happens because the single sample determines the prototype in the 1-shot task, and this uncertainty affects the classification performance.
It is worth noting that $\gamma$ only affects the behavior of the text branch, and LPN outperforms the baseline model with different $\gamma$, demonstrating the effectiveness of using class-level features as anchors in low-data scenarios.

\begin{table}[!t]
    \centering
    \caption{Effect of $\mathcal{L}_{scl}$  on \emph{mini}ImageNet.}
    \vspace{2pt}
    \label{Tab:albation_scl}
    \begin{tabular}{c | c c c c }
	\toprule
		$\beta$ & 0 & 0.1 &0.2 & 0.4\\
	\midrule
	\emph{1-shot} & 71.09 &71.82$_{{(+0.73)}}$ &71.88$_{{(+0.79)}}$ &71.99$_{{(+0.90)}}$\\

\end{tabular}
\end{table}

\subsubsection{Effect of $\mathcal{L}_{scl}$}
To address the issue of confusion that maybe arise when generating text features directly from class-level features, we employ a supervised contrastive loss $\mathcal{L}_{scl}$ on the text branch, as described in Section~\ref{sec:loss}. As listed on Table~\ref{Tab:albation_scl}, adding $\mathcal{L}_{scl}$ further improves performance by 0.90\% in 1-shot tasks with $\beta$=0.4 and $\gamma$=0.01.

\subsection{More Analysis}

\begin{table}[!t]
    \centering
    \caption{The results of LPN with different baseline models on \emph{mini}ImageNet.}
    \vspace{2pt}
    \label{Tab:Gn_LPN}

    \begin{tabular}{lccc}
    
    \toprule
    \multirow{2}{*}{Method}&\multirow{2}{*}{LPN}&\multicolumn{2}{c}{Shot}\\ 
    \cline{3-4}
    && \emph{1-shot} & \emph{5-shot}\\ 
    \midrule
    \multirow{2}{*}{ProtoNet~\cite{snell2017prototypical}} 
    &\emph{w/o}  &63.54 &	80.79 \\
    &\emph{w/} &71.99$_{{(+8.45)}}$ &	82.43$_{{(+1.64)}}$\\ 
    \midrule

    \multirow{2}{*}{DN4~\cite{li2019revisiting}} 
    &\emph{w/o}  & 62.15&	78.03 \\
    &\emph{w/} & 70.64$_{{(+8.49)}}$&	79.09$_{{(+1.06)}}$ \\ 
    \midrule
    
    \multirow{2}{*}{RelationNet~\cite{sung2018learning}} 
    &\emph{w/o}  & 61.31&	74.36 \\
    &\emph{w/} & 69.87$_{{(+8.56)}}$	&75.29$_{{(+0.93)}}$\\ 
    \bottomrule

    \end{tabular}
\end{table}

\subsubsection{Generalization of LPN}
To further investigate the effectiveness of language modality in few-shot classification, we evaluate the generalization of LPN on \emph{mini}ImageNet using other conventional metric-based methods, DN4~\cite{li2019revisiting} and RN~\cite{sung2018learning}. We take the average of local descriptor in DN4-based LPN and employ a sigmoid function in RN-based LPN to normalize the logits from the visual branch, as the text logits are computed with cosine similarity. Table~\ref{Tab:Gn_LPN} shows that LPN significantly improves performance in 1-shot tasks. Furthermore, LPN also obtains some improvement in 5-shot tasks. The results demonstrate that incorporating language information can be a promising way to improve few-shot classification performance.

% \noindent
\subsubsection{Analysis of Text Encoder}
While our LPN has shown promising results in previous settings, its performance has only been evaluated using CLIP with RN50 weights, which, like our LaGD module, is designed to leverage the visual-language relationship. Thus, it remains unclear whether the performance is influenced by the choice of text encoder. To address this, we conduct additional experiments using various pre-trained weights and other natural language processing(NLP) models while keeping the prototypical network as our baseline. 
The results are shown in Table~\ref{Tab:Gn_TE}. BERT~\cite{devlin2018bert} stands for bidirectional encoder representations from transformers. GloVe refers to the global vector of work representation. The experiments demonstrate that LPN consistently improves the baseline model, regardless of the text encoder used. Specifically, in 1-shot tasks, LPN achieved improvements of 8.45\% and 5.09\% with different CLIP weights while also performing well with BERT and GloVe, which is pre-trained without language-image paired settings. The results demonstrate that language modality effectively improves the performance of the few-shot classifier.

\begin{table}[!t]
    \centering
    \caption{The effect of text encoder on \emph{mini}ImageNet.}
    \vspace{2pt}
    \label{Tab:Gn_TE}
    \begin{tabular}{lcc}
    
    \toprule
    \multirow{2}{*}{Text Encoder}&\multicolumn{2}{c}{Shot}\\ 
    \cline{2-3}
    & \emph{1-shot} & \emph{5-shot}\\ 
    \midrule
    baseline &63.54 &	80.79\\
    \midrule
    RN50   &71.99$_{{(+8.45)}}$ &	82.43$_{{(+1.64)}}$ \\
    ViT/16 &68.63$_{{(+5.09)}}$ &81.95$_{{(+1.16)}}$ \\
    \midrule
    BERT~\cite{devlin2018bert} &65.74$_{{(+2.20)}}$ &81.94$_{{(+1.15)}}$ \\
    GloVe-300~\cite{pennington2014glove} &66.65$_{{(+3.02)}}$ &81.53$_{{(+0.74)}}$ \\
    \bottomrule
    \end{tabular}
\end{table}
% CLIP~\cite{radford2021learning}(

% \noindent
\subsubsection{Analysis of the Aggregated Logits}
To calibrate the deviation of a single modality, LPN aggregates the visual and text logits from two branches to produce the final decision. The experiments on combining the two branches are presented in Table~\ref{Tab:V_T}. $V_s$ and $V_t$ indicate the predictions generated from the visual and text branch, respectively. The plus sign refers to the post-fusion of our method. The results demonstrate that the visual branch outperforms the text branch in 5-shot tasks, benefiting from more valid support features. On the other hand, the text branch performs well in 1-shot tasks, leveraging class-level features to construct robust prototypes. However, the text branch struggles in 5-shot tasks, likely due to the lack of detailed expression of individual characteristics in the text features, which are constructed solely from class-level features rather than image descriptions. By aggregating the two logits in the post-fusion, LPN avoids the instability of a single modality and combines the strengths of both visual and text features.

\begin{table}[!t]
    \centering
    \caption{The influence of aggregated logits on \emph{mini}ImageNet. We conduct experiments on trained LPN using different logits as predictions. The baseline refers to the ProtoNet in our settings.}
    \vspace{2pt}
    \label{Tab:V_T}
    \begin{tabular}{lcc}
    
    \toprule
    \multirow{2}{*}{Logits}&\multicolumn{2}{c}{Shot}\\ 
    \cline{2-3}
    & \emph{1-shot} & \emph{5-shot}\\ 
    \midrule
    baseline  &63.54 &	80.79\\
    \midrule
    $V_s$   &61.51$_{{(-2.03)}}$ &82.37$_{{(+1.58)}}$ \\
    $V_t$    &68.71$_{{(+5.17)}}$ &77.58$_{{(-3.21)}}$ \\
    \midrule
    $V_s+V_t$  &71.99$_{{(+8.45)}}$ &	82.43$_{{(+1.64)}}$ \\
    \bottomrule
    \end{tabular}
\end{table}

\begin{figure}[!t]
    \centering
    \subfloat[5-way K-shot]{
        \includegraphics[width=0.8\linewidth]{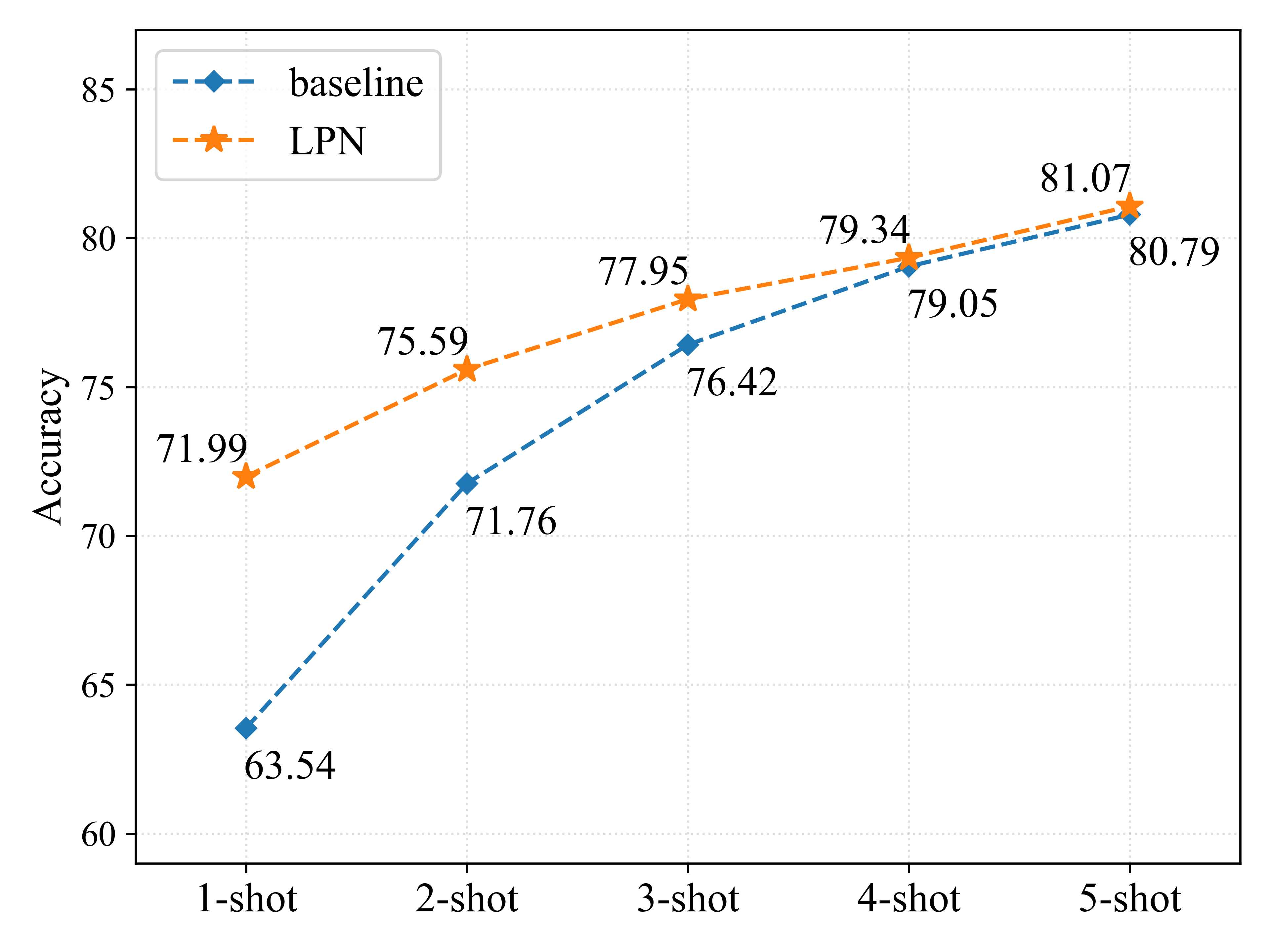}
        \label{fig:NW.Kshot}
    }
    \\
    \subfloat[N-way 1-shot]{
	\includegraphics[width=0.8\linewidth]{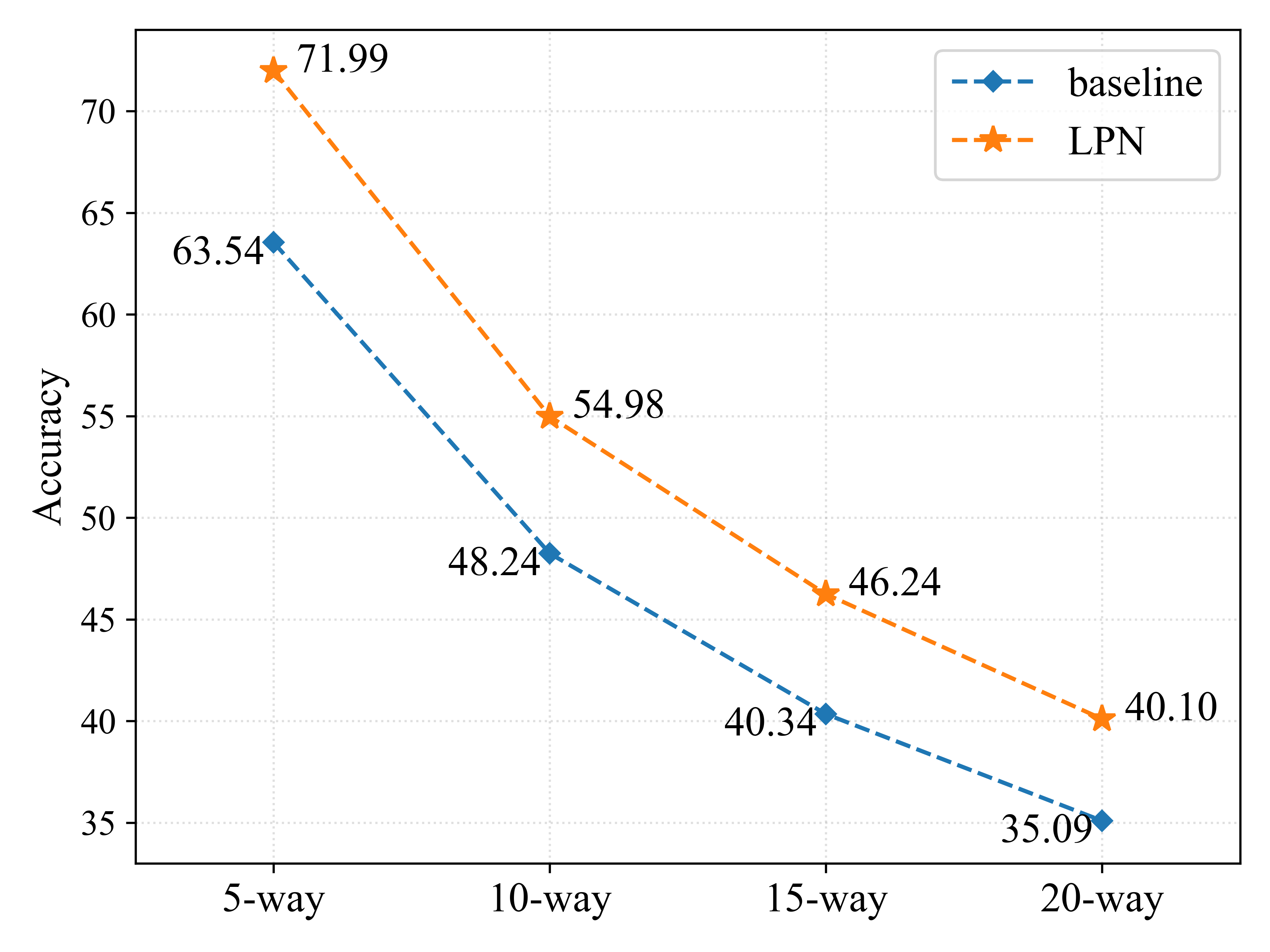}
        \label{fig:NW.Nway}
    }
    \caption{The performance of LPN on \emph{mini}ImageNet with N-way K-shot settings. The baseline refers to the results of ProtoNet in our settings.}
    \label{fig:NW}
\end{figure}

% \noindent
\subsubsection{Influence of Class Numbers and Shots in Each Task}
To further investigate the impact of language modality, we evaluate the performance of LPN with $\gamma$=0.01 in 5-way K-shot tasks and compare it to the prototypical network~\cite{snell2017prototypical}. As illustrated in Figure~\ref{fig:NW.Kshot}, LPN outperforms the baseline model in all K-shot settings. Notably, LPN performs exceptionally well with fewer support settings. 
Besides, to observe the stability of LPN, we conduct experiments on \emph{mini}ImageNet with N-way 1-shot settings. The results are shown in Figure~\ref{fig:NW.Nway}. LPN improves across a range of N-way settings, from 5 to 20. The results highlight the benefits of multi-modality in few-shot classification and provide further evidence of the effectiveness of our proposed approach.

% \noindent
\subsubsection{Visualization}

\begin{figure*}[!t]
\centering
\includegraphics[width=1.0\linewidth]{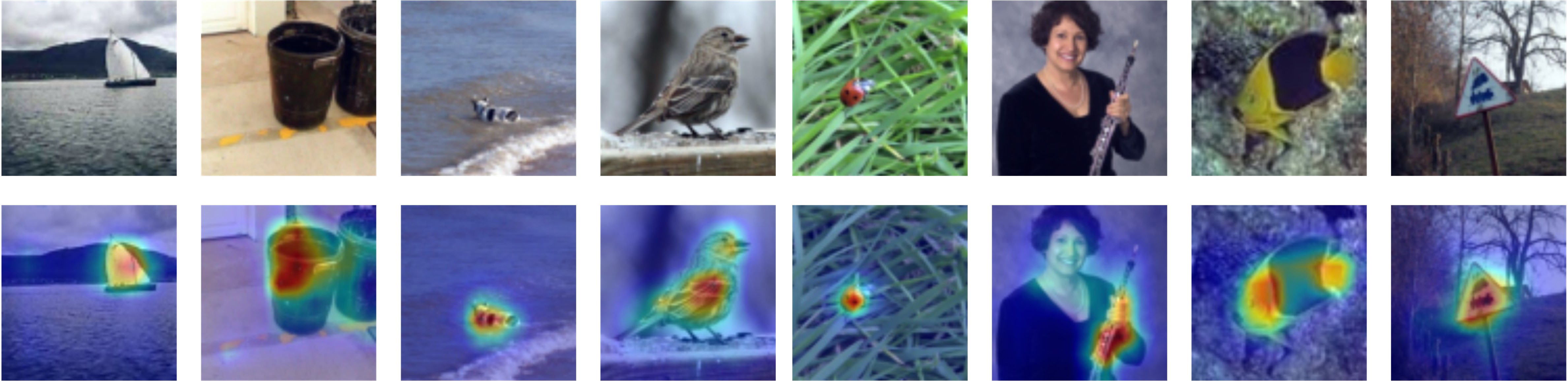}
\caption{The visualization of cross-attention in LaGD on \emph{mini}ImageNet}
\label{fig:LaGD_attn}
\end{figure*}

\begin{figure*}[!t]
\centering
\includegraphics[width=1.0\linewidth]{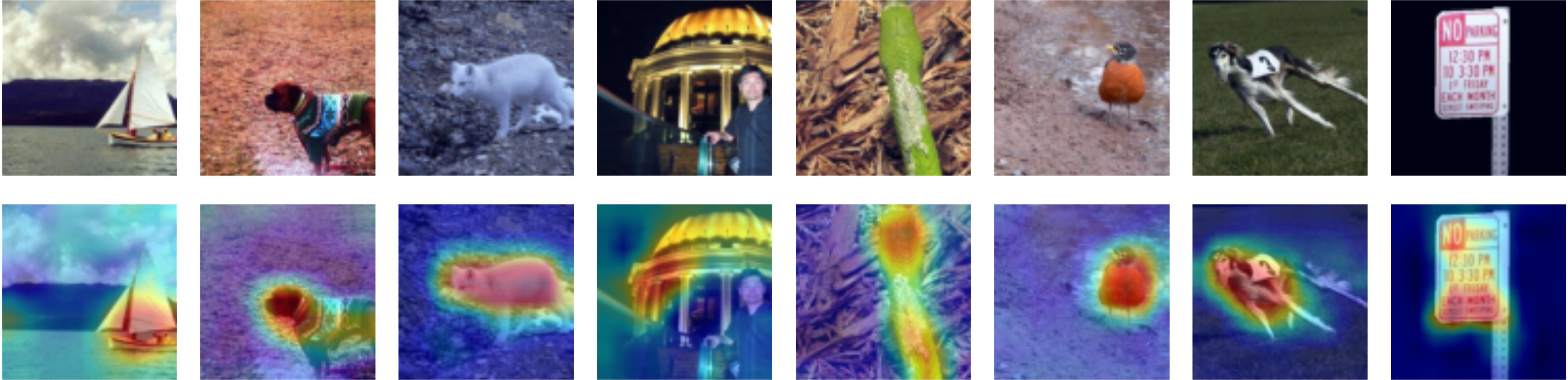}
\caption{The attention weight of LA on \emph{mini}ImageNet}
\label{fig:LA_visual}
\end{figure*}

The correlation between visual and class-level text features is calculated through the attention module, LaGD, as described in Section~\ref{sec:LaGD}. We visualize the attention mask in the first cross-attention in LaGD, designed to measure the relation between the two modalities. As illustrated in Figure~\ref{fig:LaGD_attn}, the first rows are the original input images, and the second are the corresponding heatmaps generated from the cross-attention module. Our LPN can grasp the relation between visual and class-level features through the LaGD, such as yawl and road sign in the pictures. This enables LPN to generate corresponding text features concerning image features. We simultaneously visualize the influence of LA, proposed in Section~\ref{sec:LA}, and the result is shown in Figure~\ref{fig:LA_visual}. The results indicates that using text features as convolutional kernel makes the model focus on category-related features, which contributes to model performance.

%%%%%%%%%%%%%%%%%%%%%%%%%%%%%%%%%
\section{CONCLUSIONS}\label{sec:CONCLUSION}

In this paper, we propose a language-guided prototypical network (LPN) for few-shot image classification. The proposed LPN consists of two branches. One is the visual branch, which embeds the input images and leverage class-level text features to emphasize essential features. This branch measures similarity between queries and supports using metrics. Another is the text branch, which acquires logits through two modules to promote visual features. We introduce two fusion stages to integrate the two branches. In the pre-fusion stage, the text branch leverages the language-guided decoder and a pre-trained text encoder to learn text features corresponding to the images due to the lack of text descriptions. Then, it employs the refined prototypical head to obtain more robust prototypes for text logits. In the post-fusion stage, we aggregate the visual and text logits to calibrate the two branches for final decision-making. Extensive experiments demonstrate the competitive performance of our LPN when compared to state-of-the-art methods, especially in 1-shot tasks. Furthermore, we demonstrate the versatility of LPN by successfully applying it to other conventional few-shot classifiers, highlighting the potential of combining multi-modalities in few-shot learning.

\bibliographystyle{IEEEtran}
\bibliography{ref}

% Biography text here without a photo.

% \begin{IEEEbiography}[{\includegraphics[width=1in,height=1.25in,clip,keepaspectratio]{fig1.png}}]{IEEE Publications Technology Team}
% In this paragraph you can place your educational, professional background and research and other interests.\end{IEEEbiography}

\end{document}